# Exploring Local Explanations of Nonlinear Models Using Animated Linear Projections

Nicholas Spyrison[1*], Dianne Cook[2] and Przemyslaw Biecek[3]

[1*]Faculty of Information Technology, Monash University, Wellington Rd, Clayton, VIC 3800, Australia.
[2]Dept of Econometrics and Business Statistics, Monash University, Wellington Rd, Clayton, VIC 3800, Australia.
[3]Faculty of Mathematics and Information Science, Warsaw University of Technology, Koszykowa 75, Warsaw, 00-662, Poland.

*Corresponding author(s). E-mail(s): spyrison@gmail.com;
Contributing authors: dicook@monash.edu;
przemyslaw.biecek@pw.edu.pl;

**Abstract**

The increased predictive power of machine learning models comes at the cost of increased complexity and loss of interpretability, particularly in comparison to parametric statistical models. This trade-off has led to the emergence of eXplainable AI (XAI) which provides methods, such as local explanations (LEs) and local variable attributions (LVAs), to shed light on how a model use predictors to arrive at a prediction. These provide a point estimate of the linear variable importance in the vicinity of a single observation. However, LVAs tend not to effectively handle association between predictors. To understand how the interaction between predictors affects the variable importance estimate, we can convert LVAs into linear projections and use the radial tour. This is also useful for learning how a model has made a mistake, or the effect of outliers, or the clustering of observations. The approach is illustrated with examples from categorical (penguin species, chocolate types) and quantitative (soccer/football salaries, house prices) response models. The methods are implemented in the R package cheem, available on CRAN.

# 1 Introduction

There are different reasons and purposes for fitting a model. According to the taxonomies of Breiman (2001b) and Shmueli (2010), it can be useful to group models into two types: explanatory and predictive. Explanatory modeling is used for inferential purposes, while predictive modeling focuses solely on the performance of an objective function. The intended use of the model has important implications for its selection and development. Interpretability is critical in explanatory modeling to draw meaningful inferential conclusions, such as which variables most contribute to a prediction or whether some observations are less well fit. Interpretability becomes more difficult when the model is nonlinear. Nonlinear models occur in statistical models with polynomial or interaction terms between quantitative predictors, and almost all computational models such as random forests, support vector machines, or neural networks (e.g. Breiman, 2001a; Boser et al., 1992; Anderson, 1995).

In linear models interpretation of the importance of variables is relatively straightforward, one adjusts for the covariance of multiple variables when examining the relationship with the response. The interpretation is valid for the full domain of the predictors. In nonlinear models, one needs to consider the model in small neighborhoods of the domain to make any assessment of variable importance. Even though this is difficult, it is especially important to interpret model fits as we become more dependent on nonlinear models for routine aspects of life to avoid issues described in Stahl (2021). Understanding how nonlinear models behave when usage extrapolates outside the domain of predictors, either in sub-spaces where few samples were provided in the training set, or extending outside the domain. It is especially important because nonlinear models can vary wildly and predictions can be dramatically wrong in these areas.

Explainable Artificial Intelligence (XAI) is an emerging field of research focused on methods for the interpreting of models (Adadi and Berrada, 2018; Barredo Arrieta et al., 2020). A class of techniques, called *local explanations* (LEs), provide methods to approximate linear variable importance, called local variable attributions (LVAs), at the location of each observation or the predictions at a specific point in the data domain. Because these are point-specific, it is challenging to comprehensively visualize them to understand a model. There are common approaches for visualizing high-dimensional data as a whole, but what is needed are new approaches for viewing these individual LVAs relative to the whole.

For multivariate data visualization, a *tour* (Asimov, 1985; Buja and Asimov, 1986; Lee et al., 2021) of linear data projections onto a lower-dimensional space, could be an element of XAI, complementing LVAs. Applying tours to model interpretation is recommended by Wickham et al. (2015) primarily to examine the fitted model in the space of the data. Cook et al. (2007) describe the use of tours for exploring classification boundaries and model diagnostics (Caragea et al., 2008; Lee et al., 2013; da Silva et al., 2021). There are various types of tours. In a *manual* or radial tour (Cook and Buja, 1997; Spyrison and Cook, 2020), the path of linear projections is defined by changing the contribution of a selected variable. We propose to use this to scrutinize the LVAs. This approach could be considered to be a counter-factual, what-if analysis, such as *ceteris paribus* ("other things held constant") profiles (Biecek, 2020).

The remainder of this paper is organized as follows. Section 2 covers the background of the LEs and the traditional visuals produced. Section 3 explains the tours and particularly the radial manual tour. Section 4 discusses the visual layout in the graphical user interface and how it facilitates analysis, data pre-processing, and package infrastructure. Illustrations are provided in Section 5 for a range of supervised learning tasks with categorical and quantitative response variables. These show how the LVAs can be used to get an overview of the model's use of predictors and to investigate errors in the model predictions. Section 6 concludes with a summary of the insights gained. The methods are implemented in the **R** package **cheem**.

## 2 Local Explanations

LVAs shed light on machine learning model fits by estimating linear variable importance in the vicinity of a single observation. There are many approaches for calculating LVAs. A comprehensive summary of the taxonomy of currently available methods is provided in Figure 6 by Barredo Arrieta et al. (2020). It includes a large number of model-specific explanations such as deepLIFT (Shrikumar et al., 2016, 2017), a popular recursive method for estimating importance in neural networks. There are fewer model-agnostic methods, of which LIME (Ribeiro et al., 2016) and SHaply Additive exPlanations (SHAP) (Lundberg and Lee, 2017), are popular.

These observation-level explanations are used in various ways depending on the data. In image classification, where pixels correspond to predictors, saliency maps overlay or offset a heatmap to indicate important pixels (Simonyan et al., 2014). For example, pixels corresponding to snow may be highlighted as important contributors when distinguishing if a picture contains a coyote or husky. In text analysis, word-level contextual sentiment analysis highlights the sentiment and magnitude of influential words (Vanni et al., 2018). In the case of numeric regression, they are used to explain additive contributions of variables from the model intercept to the observation's prediction (Ribeiro et al., 2016).

We will be focusing on SHAP values in this paper, but the approach is applicable to any method used to calculate the LVAs. SHAP calculates the variable contributions of one observation by examining the effect of other variables on the predictions. The term "SHAP" refers to Shapley (1953)'s method to evaluate an individual's contribution in cooperative games by assessing this player's performance in the presence or absence of other players. Strumbelj and Kononenko (2010) introduced SHAP for LEs in machine learning models. Variable importance can depend on the sequence in which variables are entered into the model fitting process, thus for any sequence we get a set of variable contribution values for a single observation. These values will add up to the difference between the fitted value for the observation, and the average fitted value for all observations. Using all possible sequences, or permutations, gives multiple values for each variable, which are averaged to get the SHAP value for an observation. It can be helpful to standardize variables prior to computing SHAP values if they have been measured on different scales.

The approach is related to partial dependence plots (for example see chapter 8 of Molnar (2022)), used to explain the effect of a variable by predicting the response for a range of values on this variable after fixing the value of all other variables to their mean. Though partial dependence plots are a global approximation of the variable importance, while SHAP is specific to one observation.

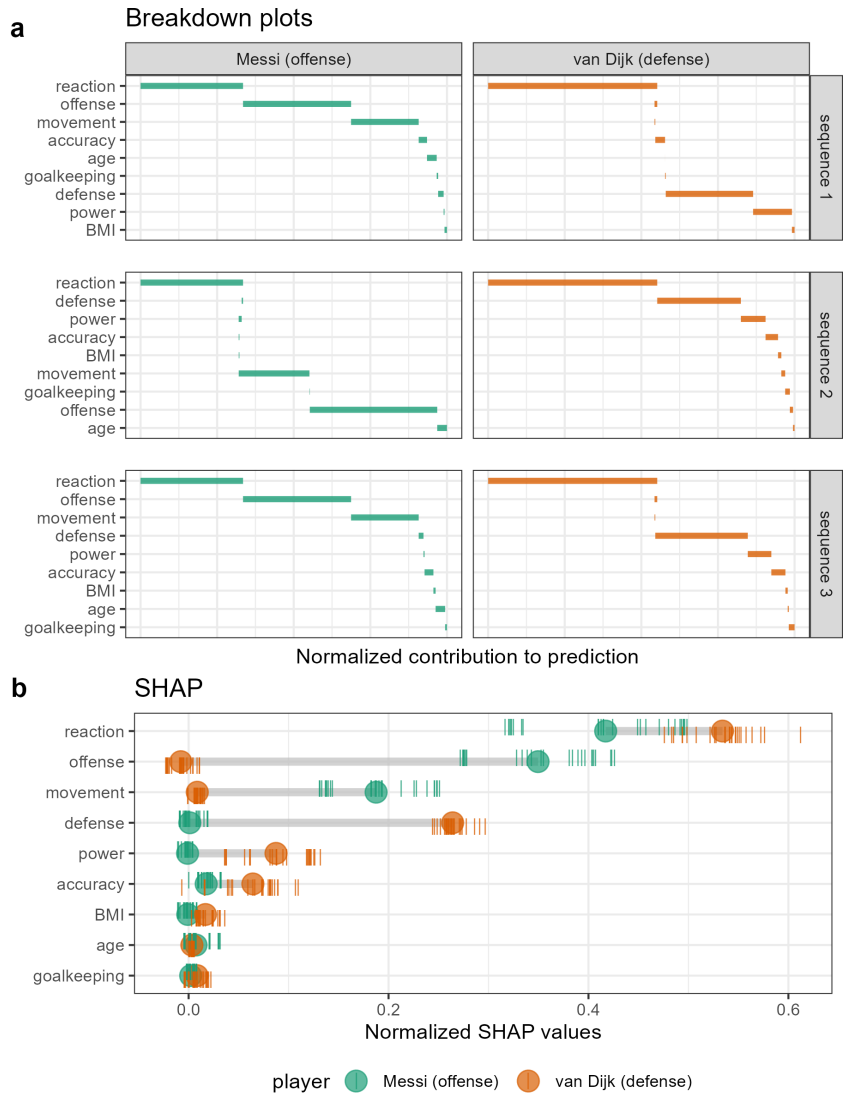

**Fig. 1** Illustration of SHAP values for a random forest model FIFA 2020 player wages from nine skill predictors. A star offensive and defensive player are compared, L. Messi and V. van Dijk, respectively. Panel (a) shows breakdown plots of three sequences of the variables. The sequence of the variables impacts the magnitude of their attribution. Panel (b) shows the distribution of attribution for each variable across 25 sequences of predictors, with the mean displayed as a dot for each player. Reaction skills are important for both players. Offense and movement are important for Messi but not van Dijk, and conversely, defense and power are important for van Dijk but not Messi.

We use 2020 season FIFA data (Leone, 2020) to illustrate SHAP following the procedures described in Biecek and Burzykowski (2021). There are 5000 observations of nine predictor variables measuring players' skills and one response variable, wages (in euros). A random forest model is fit regressing players' wages on the skill variables. In this illustration in Figure 1 the SHAP values are compared for a star offensive player (L. Messi) and a prominent defensive player (V. van Dijk). We are interested in knowing how the skill variables locally contribute to the wage prediction of each player. A difference in the attribution of the variable importance across the two positions of the players can be expected. This would be interpreted as how a player's salary depends on which combination of skills. Panel (a) is a version of a breakdown plot (Gosiewska and Biecek, 2019) where just three sequences of variables are shown, for two observations. A breakdown plot shows the absolute values of the variable attribution for an observation, usually sorted from the highest value to the lowest. There is no scale on the horizontal axis here because values are considered relative to each other. Here we can see how the variable contribution can change depending on sequence, relative to both players. (Note that the order of the variables is different in each plot because they have been sorted by the biggest average contribution across both players.) For all sequences, and for both players `reaction` has the strongest contribution, with perhaps more importance for the defensive player. Then it differs by player: for Messi `offense` and `movement` have the strongest contributions, and for van Dijk it is `defense` and `power`, regardless of the variable sequence.

Panel (b) shows the differences in the player's median values (large dots) for 25 such sequences (tick marks). We can see that the wage predictions for the two players come from different combinations of skill sets, as might be expected for players whose value on the team depends on their offensive or defensive prowess. It is also interesting to see from the distribution of values across the different sequences of variables, that there is some multimodality. For example, look at the SHAP values for `reaction` for Messi, and in some sequences, reaction has a much lower contribution than others. This suggests that other variables (`offense`, `movement` probably) can substitute for `reaction` in the wage prediction.

This can also be considered similar to examining the coefficients from all subsets regression, as described in Wickham et al. (2015). Various models that are similarly good might use different combinations of the variables. Examining the coefficients from multiple models helps to understand the relative importance of each variable in the context of all other variables. This is similar to the approach here with SHAP values, that by examining the variation in values across different permutations of variables, we can gain more understanding of the relationship between the response and predictors.

For the application, we use *tree SHAP*, a variant of SHAP that enjoys a lower computational complexity (Lundberg et al., 2018). Instead of aggregating over sequences of the variables, tree SHAP calculates observation-level variable importance by exploring the structure of the decision trees. Tree SHAP is only compatible with tree-based models. so random forests are used for illustration.

There are numerous R packages currently available on CRAN that provide functions for computing SHAP and other LVA values, including `treeshap` (Kominsarczyk et al., 2023), `fastshap` (Greenwell, 2023), `kernelshap` (Mayer and Watson, 2023),

`shapr` (Sellereite et al., 2023), `shapviz` (Mayer, 2023b), `PPtreeregViz` (Lee and Cho, 2022), `ExplainPrediction` (Robnik-Sikonja, 2018), `flashlight` (Mayer, 2023a), and the package `DALEX` has many resources (Biecek, 2018). Molnar (2022) provides good explanations of the different methods and how to apply them to different models.

# 3 Tours and the Radial Tour

A *tour* enables the viewing of high-dimensional data by animating many linear projections with small incremental changes. It is achieved by following a path of linear projections (bases) of high-dimensional space. One key variable of the tour is the object permanence of the data points; one can track the relative change of observations in time and gain information about the relationships between points across multiple variables. There are various types of tours that are distinguished by how the paths are generated (Lee et al., 2021; Cook et al., 2008).

The manual tour (Cook and Buja, 1997) defines its path by changing a selected variable's contribution to a basis to allow the variable to contribute more or less to the projection. The requirement constrains the contribution of all other variables that a basis needs to be orthonormal (columns correspond to vectors, with unit length, and orthogonal to each other). The manual tour is primarily used to assess the importance of a variable to the structure visible in a projection. It also lends itself to pre-computation queued in advance or computed on the fly for human-in-the-loop analysis (Karwowski, 2006).

A version of the manual tour called a *radial tour* is implemented in Spyrison and Cook (2020) and forms the basis of this new work. In a radial tour, the selected variable can change its magnitude of contribution but not its angle; it must move along the direction of its original contribution. The implementation allows for pre-computation and interactive re-calculation to focus on a different variable. In this work, the radial tour allows us to explore the sensitivity of LVA to the prediction of a model.

# 4 The Cheem Viewer

To explore the LVAs, coordinated views (Roberts, 2007) (also known as ensemble graphics, Unwin and Valero-Mora, 2018) are provided in the *cheem viewer* application. There are two primary plots: the **global view** to give the context of all of the SHAP values and the **radial tour view** to explore the LVAs with user-controlled rotation. There are numerous user inputs, including variable selection for the radial tour and observation selection for making comparisons. There are different plots used for the categorical and quantitative responses. Figures 3 and 4 are screenshots showing the cheem viewer for the two primary tasks: classification (categorical response) and regression (quantitative response).

## 4.1 Global View

The global view provides context for all observations and facilitates the exploration of the separability of the data and attribution spaces. The attribution space refers to the SHAP values for each observation. These spaces both have dimensionality $n \times p$, where $n$ is the number of observations and $p$ is the number of variables.

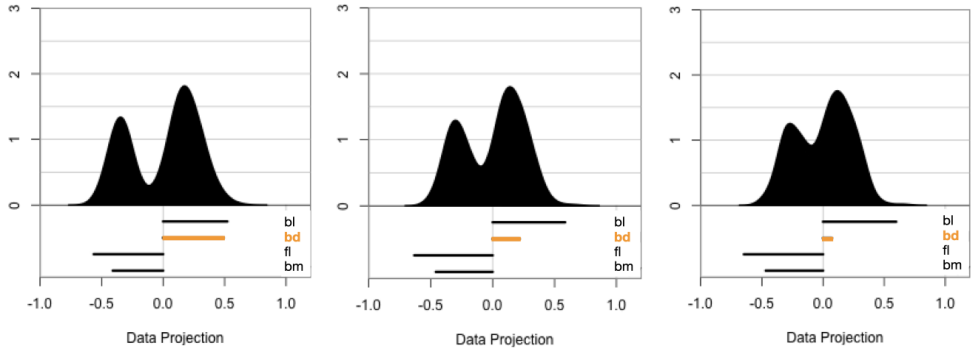

**Fig. 2** The radial tour allows the user to remove a variable from a projection, to examine the importance of this variable to the structure in the plot. Here we have a 1D projection of the penguins data displayed as a density plot. The line segments on the bottom correspond to the coefficients of the variables making up the projection. The structure in the plot is bimodality (left), and the importance of the variable `bd` is being explored. As this variable contribution is reduced in the plot (middle, right) we can see that the bimodality decreases. Thus `bd` is an important variable contributing to the bimodal structure.

The visualization is composed of the first two principal components of the data (left) and the attribution (middle) spaces. These single 2D projections will not reveal all of the structure of higher-dimensional space, but they are helpful visual summaries. In addition, a plot of the observed against predicted response values is also provided (Figures 3c, 4b) to help identify observations poorly predicted by the model. For classification tasks, color indicates the predicted class and misclassified observations are circled in red. Linked brushing between the plots is provided (click and drag), and a tabular display of selected points helps to facilitate the exploration of the spaces and the model (shown in Figure 4d).

While the comparison of these spaces is interesting, the primary purpose of the global view is to enable the selection of particular observations to explore in detail. We have designed it to enable a comparison between an observation that is interesting in some way, perhaps misclassified, or poorly predicted, relative to an observation with similar predictor values but a more expected prediction. For brevity, we call the interesting observation the primary investigation (PI), and the other is the comparison investigation (CI). These observations are highlighted as an asterisk and ×, respectively.

### 4.2 Radial Tour

The radial tour is used to explore how the SHAP value of a variable relates to it's effect on the predicted value. In a similar way as explained in Section 3, where the radial tour is used to understand a variable's contribution to cluster structure, for model prediction explanations, the radial tour is used to understand a variable's contribution to the observation's predicted value. By altering the contribution using the radial tour, we see how the predicted value might change. If a small change in the variable contribution results in a big change in predicted value, then this variable substantially explains the model prediction. The SHAP values are estimates of the local importance, and provide a good starting place from which to begin a radial tour. They can be misleading, and the radial tour can help to assess the strength of the explanatory power of the SHAP value. Because the SHAP values are local, using linear projections to explore a local neighborhood of a nonlinear model is reasonable.

There are two plots in this part of the interface. The first (Figures 3e and 4e) is a display of the SHAP values for all observations. This will generally give the global view of variables important for the fit as a whole, but it will also highlight observations that have different patterns. The second plot is the radial tour, which for classification is a density plot of a 1D projection (Figure 3f), and for regression are scatterplots of the observed response values, and residuals, against a 1D projection (Figure 4f).

The LVAs for all observations are normalized (sum of squares equals 1), and thus, the relative importance of variables can be compared across all observations. These are depicted as a vertical parallel coordinate plot (Ocagne, 1885). (The SHAP values of the PI and CI are shown as dashed and dotted lines, respectively.) One should obtain a sense of the overall importance of variables from this plot. The more important variables will have larger values, and in the case of classification tasks variables that have different magnitudes for different classes are more globally important. For example, Figure 3e suggests that `bl` is important for distinguishing the green class from the other two. For regression, one might generally observe which variables have low values for all observations (not important). For example, `BMI` and `pwr` in Figure 4e, have a range of high and low values (e.g., `off`, `def`), suggesting they are important for some observations and not important for others.

A bar chart is overlaid to represent the projection shown in the radial tour on the right. It starts from the SHAP values of the PI, but if the user changes the projection the length of these bars will reflect this change. By scaling the SHAP value it becomes an (attribution) projection.

The attribution projection of the PI is the initial 1D basis in a radial tour, displayed as a density plot for a categorical response (Figure 3f) and as scatterplots for a quantitative response (Figure 4f). The PI and CI are indicated by vertical dashed and dotted lines, respectively. The radial tour varies the contribution of the selected variable. This is viewed as an animation of the projections from many intermediate bases. Doing so tests the sensitivity of structure (class separation or strength of relationship) to the variable's contribution. The CI attribution of the CI does not impact the bases but it highlighted from context. For classification, if the separation between classes diminishes when the variable contribution is reduced, this suggests that the variable is important for class separation. For regression, if the relationship scatterplot weakens when the variable contribution is reduced, indicating that the variable is important for accurately predicting the response.

The purpose of using both the PI and CI when using the radial tour is comparison. Remember the CI is a representative individual with an expected prediction ( correct class or small residual) and the PI is a particularly interesting individual with a less expected prediction. The radial tour would start from the attribution projection corresponding to the SHAP values of the PI, and vary the contribution of a variable where the SHAP values differ from those of the CI. The goal is then to examine how the model prediction would change for the PI if the variable contribution changed, to be more similar to that of the CI.

### 4.3 Classification Task

Selecting a misclassified observation as PI and a correctly classified point nearby in data space as CI makes it easier to examine the variables most responsible for the error. The global view (Figure 3c) displays the model confusion matrix. The radial tour is 1D and displays as density where color indicates class. An animation slider enables users to vary the contribution of variables to explore the sensitivity of the separation to that variable.

### 4.4 Regression Task

Selecting an inaccurately predicted observation as PI and an accurately predicted observation with similar variable values as CI is a helpful way to understand how the model is failing or not. The global view (Figure 4a) shows a scatterplot of the observed vs predicted values, which should exhibit a strong relationship if the model is a good fit. The points can be colored by a statistic, residual, a measure of outlyingness (log Mahalanobis distance), or correlation to aid in understanding the structure identified in these spaces.

In the radial tour view, the observed response and the residuals (vertical) are plotted against the attribution projection of the PI (horizontal). The attribution projection can be interpreted similarly to the predicted value from the global view plot. It represents a linear combination of the variables, and a good fit would be indicated when there is a strong relationship with the observed values. This can be viewed as a local linear approximation if the fitted model is nonlinear. As the contribution of a variable is varied, if the value of the PI does not change much, it would indicate that the prediction for this observation is NOT sensitive to that variable. Conversely, if the predicted value varies substantially, the prediction is very sensitive to that variable, suggesting that the variable is very important for the PI's prediction.

### 4.5 Interactive Variables

The application has several reactive inputs that affect the data used, aesthetic display, and tour manipulation. These reactive inputs make the software flexible and extensible (Figure 3a & d). The application also has more exploratory interactions to help link points across displays, reveal structures found in different spaces, and access the original data.

A tooltip displays the observation number/name and classification information while the cursor hovers over a point. Linked brushing allows the selection of points (left click and drag) where those points will be highlighted across plots (Figure 4a & b). The information corresponding to the selected points is populated on a dynamic

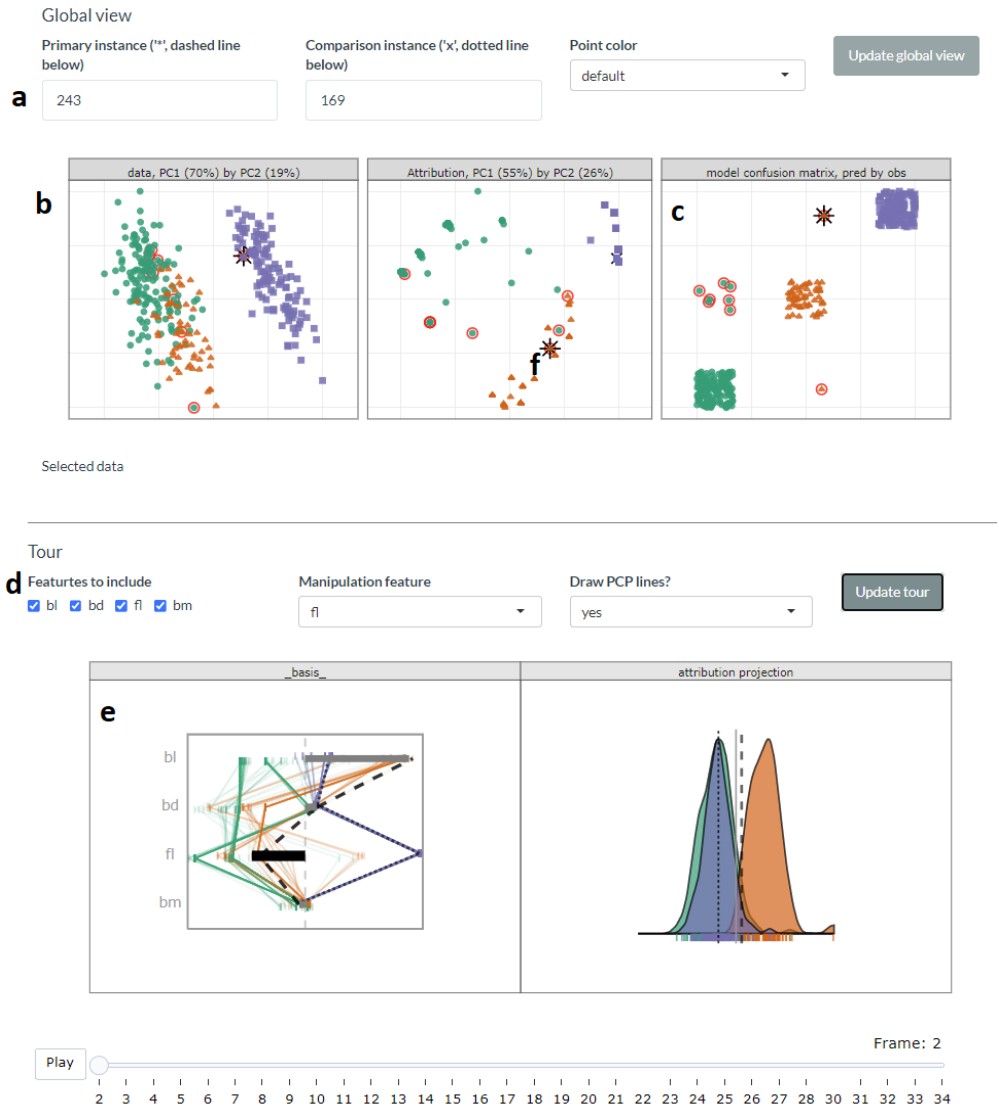

**Fig. 3** Overview of the cheem viewer for classification tasks (categorical response). Global view inputs, (a), set the PI, CI, and color statistic. Global view, (b) PC1 by PC2 approximations of the data- and attribution-space. (c) prediction by observed $y$ (visual of the confusion matrix for classification tasks). Points are colored by predicted class, and red circles indicate misclassified observations. Radial tour inputs (d) select variables to include and which variable is changed in the tour. (e) shows a parallel coordinate display of the distribution of the variable attributions while bars depict contribution for the current basis. The black bar is the variable being changed in the radial tour. Panel (f) is the resulting data projection indicated as density in the classification case.

table (Figure 4d). These interactions aid the exploration of the spaces and, finally, the identification of primary and comparison observations.

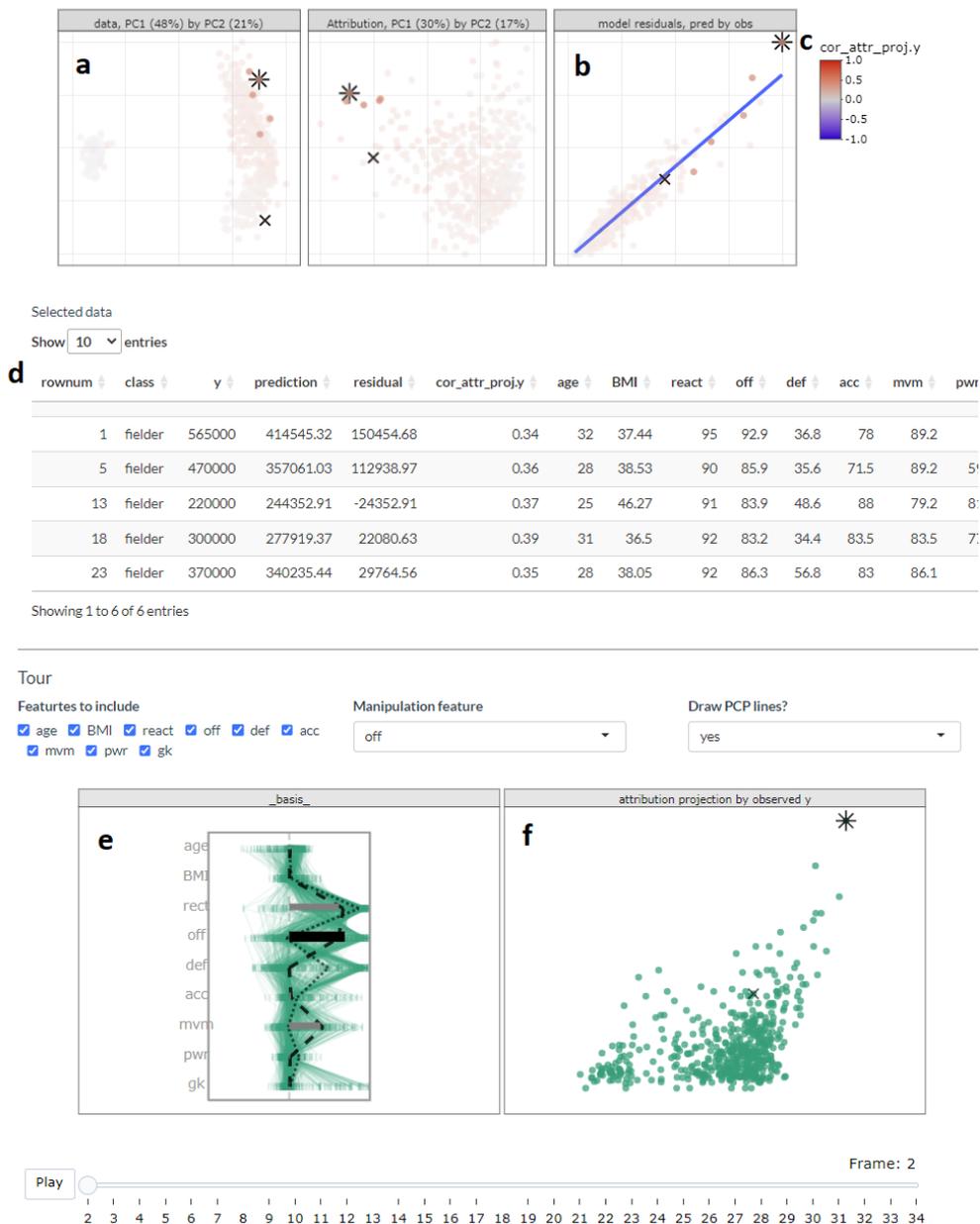

**Fig. 4** Overview of the cheem viewer for regression tasks (quantitative response) and illustration of interactive variables. Panel (a) PCA of the data- and attributions- spaces and the (b) observed vs predicted values. Four selected points are highlighted in the PC spaces and tabularly displayed. Coloring on a statistic (c) highlights the structure organized in the attribution space. Interactive tabular display (d) populates when observations are selected. Contribution of the 1D basis affecting the horizontal position (e) parallel coordinate display of the variable attribution from all observations, and horizontal bars show the contribution to the current basis. Regression projection (f) uses the same horizontal projection and fixes the vertical positions to the observed $y$ and residuals (middle and right).

### 4.6 Preprocessing

It is vital to mitigate the render time of visuals, especially when users may want to iterate many explorations. All computational operations should be prepared before run time. The work remaining when an application is run solely reacts to inputs and rendering visuals and tables. Below discusses the steps and details of the reprocessing.

- **Data:** predictors and response are unscaled complete numerical matrix. Most models and local explanations are scale-invariant. Keep the normality assumptions of the model in mind.
- **Model:** any model and compatible explanation could be explored with this method. Currently, random forest models are applied via the package **randomForest** (Liaw and Wiener, 2002), compatibility tree SHAP. Modest hyperparameters are used. Namely, classification models use 125 trees, number of variables at each split (mtry) of $\sqrt{p}$, and minimum terminal node size of $max(1, n/500)$. While regression models use 125 tree, $p/3$ variables at split, and $max(5, n/500)$ minimum terminal node size.
- **Local explanation:** Tree SHAP is calculated for *each* observation using the package **treeshap** (Kominsarczyk et al., 2023). We opt to find the attribution of each observation in the training data and not fit to fit variable interactions.
- **Cheem viewer:** after the model and full explanation space are calculated, each variable is scaled by standard deviations away from the mean to achieve common support for visuals. Statistics for mapping to color are computed on the scaled spaces.

The time to preprocess the data will vary significantly with the complexity of the model and the LE. For reference, the FIFA data contained 5000 observations of nine explanatory variables that took 2.5 seconds to fit a random forest model of modest hyperparameters. Extracting the tree SHAP values of each observation took 270 seconds in total. PCA and statistics of the variables and attributions took 2.8 seconds. These run times were from a non-parallelized session on a modern laptop, but suffice it to say that most of the time will be spent on the LVA. An increase in model complexity or data dimensionality will quickly become an obstacle. Its reduced computational complexity makes tree SHAP an excellent candidate to start. Alternatively, some package and methods use approximate calculations of LEs, such as **fastshap** Greenwell (2020).

## 5 Case Studies

To illustrate the cheem method it is applied to modern data sets, two classification examples and then two of regression.

### 5.1 Palmer Penguins, Species Classification

The Palmer penguins data (Gorman et al., 2014; Horst et al., 2020) was collected on three species of penguins foraging near Palmer Station, Antarctica. The data is publicly available to substitute for the overly-used iris data and is quite similar in form. After removing incomplete observations, there are 333 observations of four physical measurements, bill length (`bl`), bill depth (`bd`), flipper length (`fl`), and body mass (`bm`) for this illustration. A random forest model was fit with species as the response variable.

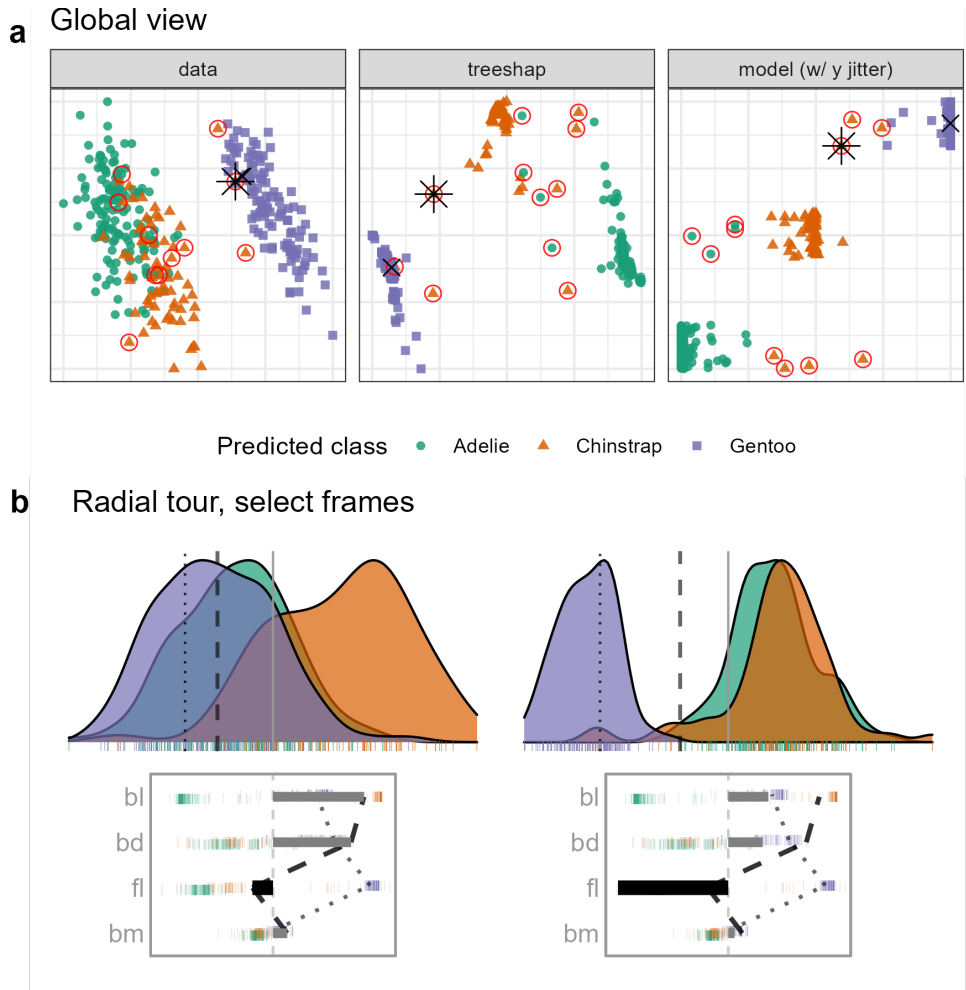

**Fig. 5** Examining the SHAP values for a random forest model classifying Palmer penguin species. The PI is a Gentoo (purple) penguin that is misclassified as a Chinstrap (orange), marked as an asterisk in (a) and the dashed vertical line in (b). The radial view shows varying the contribution of 'fl' from the initial attribution projection (b, left), which produces a linear combination where the PI is more probably (higher density value) a Chinstrap than a Gentoo (b, right). (The animation of the radial tour is at https://vimeo.com/666431172.)

Figure 5 shows plots from the cheem viewer for exploring the random forest model on the penguins data. Panel (a) shows the global view, and panel (b) shows several 1D projections generated with the radial tour. Penguin 243, a Gentoo (purple), is the PI because it has been misclassified as a Chinstrap (orange).

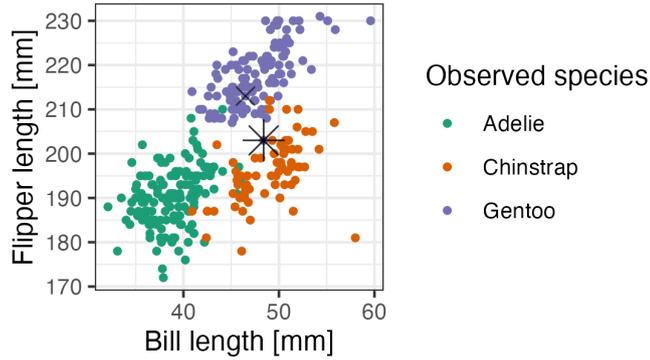

**Fig. 6** Checking what is learned from the cheem viewer. This is a plot of flipper length ('fl') and bill length ('bl'), where an asterisk highlights the PI. A Gentoo (purple) misclassified as a Chinstrap (orange). The PI has an unusually small 'fl' length which is why it is confused with a Chinstrap.

There is more separation visible in the attribution space than in the data space, as would be expected. The predicted vs observed plot reveals a handful of misclassified observations. A Gentoo which has been wrongly labeled as a Chinstrap is selected for illustration. The PI is a misclassified point (represented by the asterisk in the global view and a dashed vertical line in the tour view). The CI is a correctly classified point (represented by an × and a vertical dotted line).

The radial tour is used here to examine which variable most contributed to the incorrect classification of the PI, to understand why the model was prediction differed from that of the CI. It starts from the attribution projection of the misclassified observation (b, left). The important variables identified by SHAP in the (wrong) prediction for this observation are mostly `bl` and `bd` with small contributions of `fl` and `bm`. This projection is a view where the Gentoo (purple) looks much more likely for this observation than Chinstrap. That is, this combination of variables is not particularly useful because the PI looks very much like other Gentoo penguins. The radial tour is used to vary the contribution of flipper length (`fl`) to explore this. (In our exploration, this was the third variable explored. It is typically helpful to explore the variables with more significant contributions, here `bl` and `bd`. Still, when doing this, nothing was revealed about how the PI differed from other Gentoos). On varying `fl`, as it contributes increasingly to the projection (b, right), more and more, this penguin looks like a Chinstrap. This suggests that `fl` should be considered an important variable for explaining the (wrong) prediction.

Figure 6 confirms that flipper length (`fl`) is vital for the confusion of the PI as a Chinstrap. Here, flipper length and body length are plotted, and the PI can be seen to be closer to the Chinstrap group in these two variables, mainly because it has an unusually low value of flipper length relative to other Gentoos. From this view, it makes sense that it is a hard observation to account for, as decision trees can only partition only vertical and horizontal lines.

## 5.2 Chocolates, Milk/Dark Classification

The chocolates data set consists of 88 observations of ten nutritional measurements determined from their labels and labeled as either milk or dark. Dark chocolate is considered healthier than milk. Students collected the data during the Iowa State University class STAT503 from nutritional information on the manufacturer's websites and were normalized to 100g equivalents. The data is available in the **cheem** package. A random forest model is used for the classification of chocolate types.

It could be interesting to examine the nutritional properties of any dark chocolates that have been misclassified as milk. A reason to do this is that a dark chocolate, nutritionally more like milk should not be considered a healthy alternative. It is interesting to explore which nutritional variables contribute most to the misclassification.

This type of exploration is shown in Figure 7, where a chocolate labeled dark but predicted to be milk is chosen as the PI (observation 22). It is compared with a CI that is a correctly classified dark chocolate (observation 7). The PCA plot and the tree SHAP PCA plots (a) show a big difference between the two chocolate types but with confusion for a handful of observations. The misclassifications are more apparent in the observed vs predicted plot and can be seen to be mistaken in both ways: milk to dark and dark to milk.

The attribution projection for chocolate 22 suggests that Fiber, Sugars, and Calories are most responsible for its incorrect prediction. The way to read this plot is to see that Fiber has a large negative value while Sugars and Calories have reasonably large positive values. In the density plot, observations on the very left of the display would have high values of Fiber (matching the negative projection coefficient) and low values of Sugars and Calories. The opposite would be interpreting a point with high values in this plot. The dark chocolates (orange) are primarily on the left, and this is a reason why they are considered to be healthier: high fiber and low sugar. The density of milk chocolates is further to the right, indicating that they generally have low fiber and high sugar.

The PI (dashed line) can be viewed against the CI (dotted line). Now, one needs to pay attention to the parallel coordinate plot of the SHAP values, which are local to a particular observation, and the density plot, which is the same projection of all observations as specified by the SHAP values of the PI. The variable contribution of the two different predictions can be quickly compared in the parallel coordinate plot. The PI differs from the comparison primarily on the Fiber variable, which suggests that this is the reason for the incorrect prediction.

From the density plot, which is the attribution projection corresponding to the PI, both observations are more like dark chocolates. Using the radial tour to vary the contribution of Sugars, results in it being removed and replaced by Fiber, and reason for the wrong classification becomes apparent. In this 1D projection observation 22 is more similar to milk chocolates, suggests that Fiber is the culprit for the model mistakenly seeing it as a milk chocolate.

It would also be interesting to explore an inverse misclassification, where a milk chocolate is misclassified as a dark chocolate. Chocolate 84 is selected and is compared with a correctly predicted milk chocolate (observation 71). The corresponding global view and radial tour frames are shown in Figure 8.

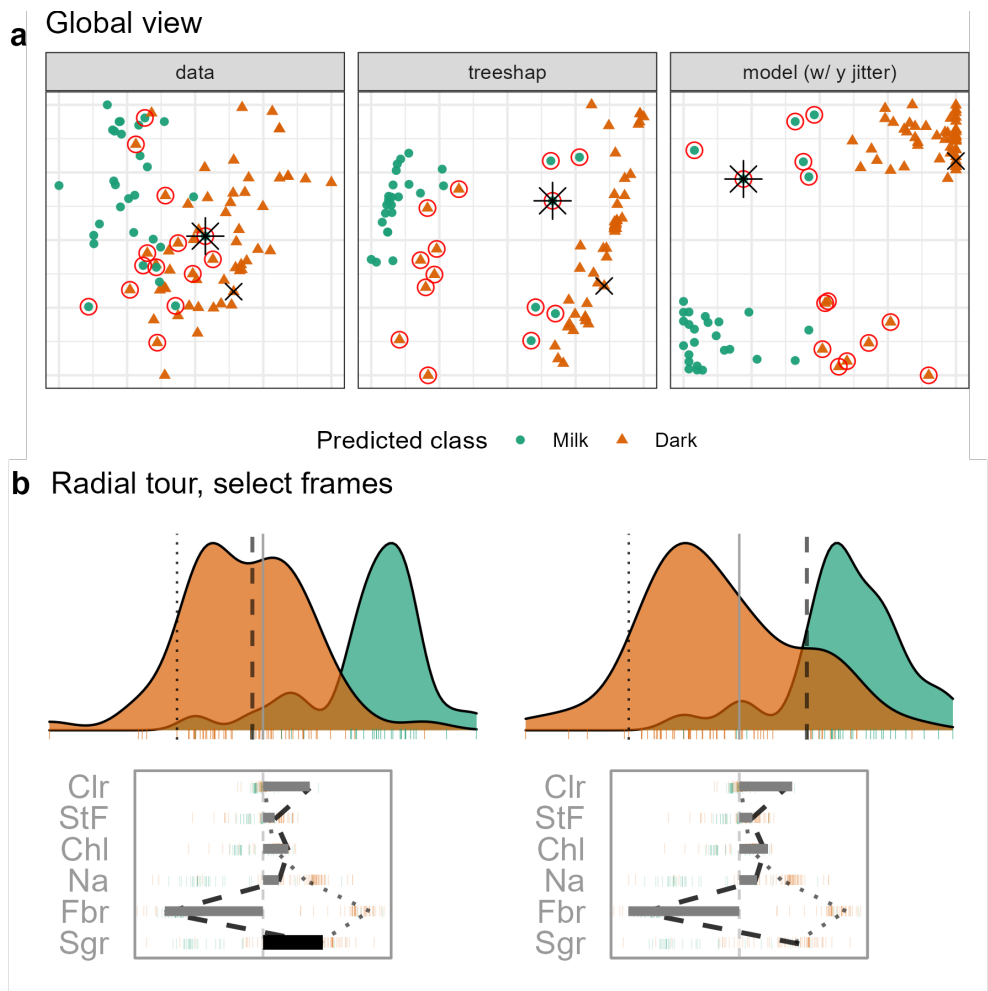

**Fig. 7** Examining the LVA for a PI which is dark (orange) chocolate incorrectly predicted to be milk (green). From the attribution projection, this chocolate correctly looks more like dark than milk, which suggests that the LVA does not help understand the prediction for this observation. So, the contribution of Sugar is varied—reducing it corresponds primarily with an increased magnitude from Fiber. When Sugar is zero, Fiber contributes strongly toward the left. In this view, the PI is closer to the bulk of the milk chocolates, suggesting that the prediction put a lot of importance on Fiber. This chocolate is a rare dark chocolate without any Fiber leading to it being mistaken for a milk chocolate. (A video of the tour animation can be found at https://vimeo.com/666431143.)

Comparing the attributions of the PI and the CI, large differences in the values of Sodium and Fiber can be seen. The contribution of Sodium is selected to be varied in the radial tour. From the density plot of the initial attribution projection, the PI is equally likely to be milk or dark dark chocolate. When the contribution of Sodium is increased, the balance shifts, and the PI is more likely to be correctly considered to be a milk chocolate. This supports that the model prediction was erroneous because it didn't adequately consider the value of Sodium in making the prediction.

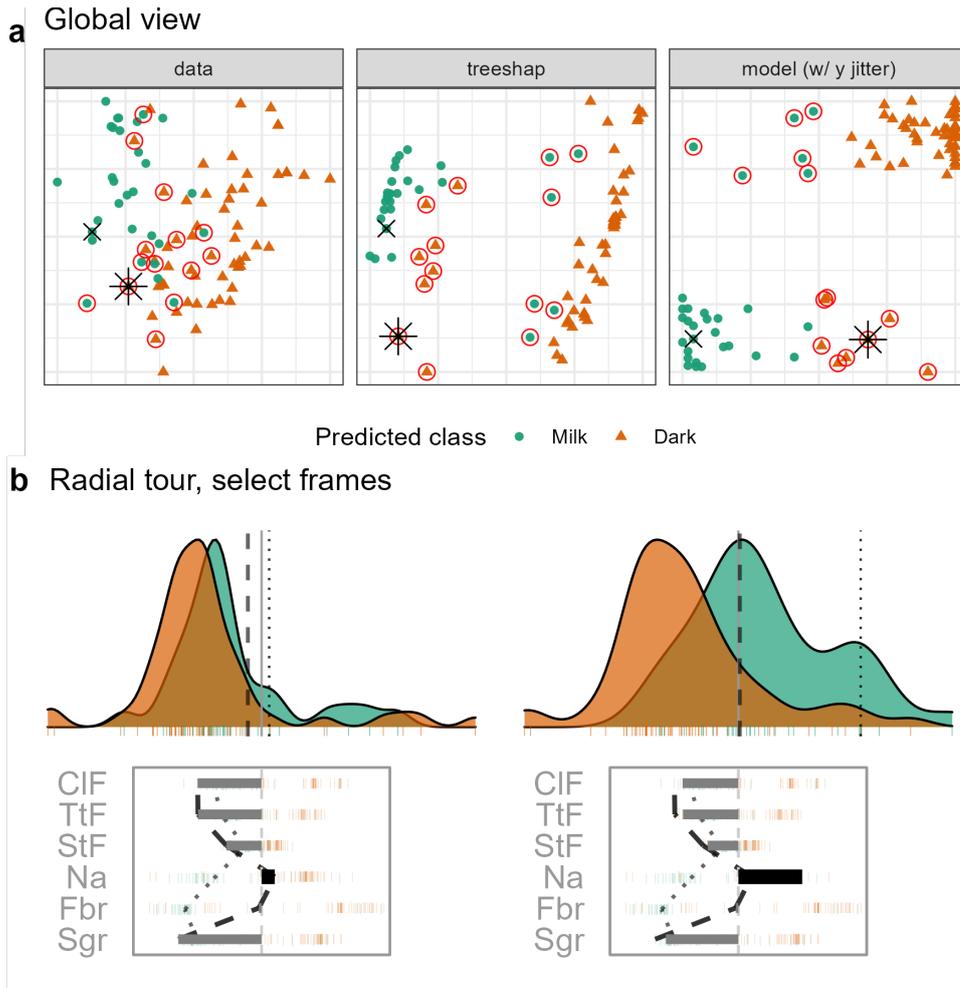

**Fig. 8** Examining the LVA for a PI which is a milk (green) chocolate incorrectly predicted to be a dark (orange). From the density plot of the attribution projection the PI could equally likely be milk or dark, where as the CI is more definitely milk. Sodium and Fiber have the largest differences in attributed variable importance, with values lose to zero, instead of large negative values like other milk chocolates. The lack of importance attributed to these variables is suspected of contributing to the mistake. When the contribution of Sodium is changedr, we see if the model had used a larger contribution of Sodium to make the prediction the PI would have likely been predicted to be a milk chocolate. (A video of the tour animation can be found at https://vimeo.com/666431148.)

### 5.3 FIFA, Wage Regression

The 2020 season FIFA data (Leone, 2020; Biecek, 2018) contains many skill measurements of soccer/football players and wage information. Nine higher-level skill groupings were identified and aggregated from highly correlated variables. A random forest model is fit from these predictors, regressing on player wages [2020 euros]. The model was fit from 5000 observations before being thinned to 500 players to mitigate occlusion and render time. Continuing from the information in Section 2, we are interested to see the difference in attribution based on what is known about different players, that is a leading offensive fielder (L. Messi) as compared with a top defensive fielder (V. van Dijk). (These same observations were shown in Figure 1.) With the radial tour we can explore how these players wages might be predicted if their skill sets were different.

Figure 9 tests the support of the LVA for the PI (Messi). The contribution from `def` is varied in the radial tour, in contrast to offensive skills (`off`). As the contribution of defensive skills increases, Messi's wage plummets. Messi's predicted wage would be much lower defensive skills played a larger role in the prediction - the model prediction reinforces that he is clearly not getting paid for his ability to defend.

Although we don't show it here, offensive and reaction (`rct`) skills are both crucial to explaining the star offensive player's predicted wage. If the contribution of either is changed, the other substitutes! That is, rotating one variable out, results in the other rotating in, when the radial tour is used, and the wage value does not change, remaining in a far-right location in the plot. Some change in predicted wage is seen if instead the contribution of a variable with low importance is varied.

### 5.4 Ames Housing, Sales Price Regression

Ames housing data (De Cock, 2011) was subset to North Ames, with 338 house sales. A random forest model was fit, predicting the sale price [USD] from the property variables: Lot Area (`LtA`), Overall Quality (`Qlt`), Year the house was Built (`YrB`), Living Area (`LvA`), number of Bathrooms (`Bth`), number of Bedrooms (`Bdr`), the total number of Rooms (`Rms`), Year the Garage was Built (`GYB`), and Garage Area (`GrA`). Using interactions with the global view, a house with an extreme negative residual and an accurate observation with a similar prediction is selected.

Figure 10 illustrates the exploration of the model predictions for the house sale 74 (PI), which is under-valued by the model. The CI has a similar predicted price though the prediction was accurate. The SHAP values for the PI and CI have very different values of Lot Area. The attribution projection would give the PI a higher value than the CI, suggesting that the Lot Area value is important for the predicted value of the PI but not for that of the CI. As the contribution of Lot Area is decreased in the radial tour, the predict value of PI decreases while the CI increases. This is quite interesting, that the SHAP value picks up the importance of Lot Area. And it appears that the model does not adequately use this variable. For the attribution projection, with a large contribution from Lot Area, the PI is better predicted than in the model, and would have a smaller residual.

## 6 Discussion

There is a clear need to provide more tools interpret black box models. Techniques such as SHAP, LIME, Break-down calculate LEs for each observation in the data. They estimate how important variables are for the model's prediction of a single observation.

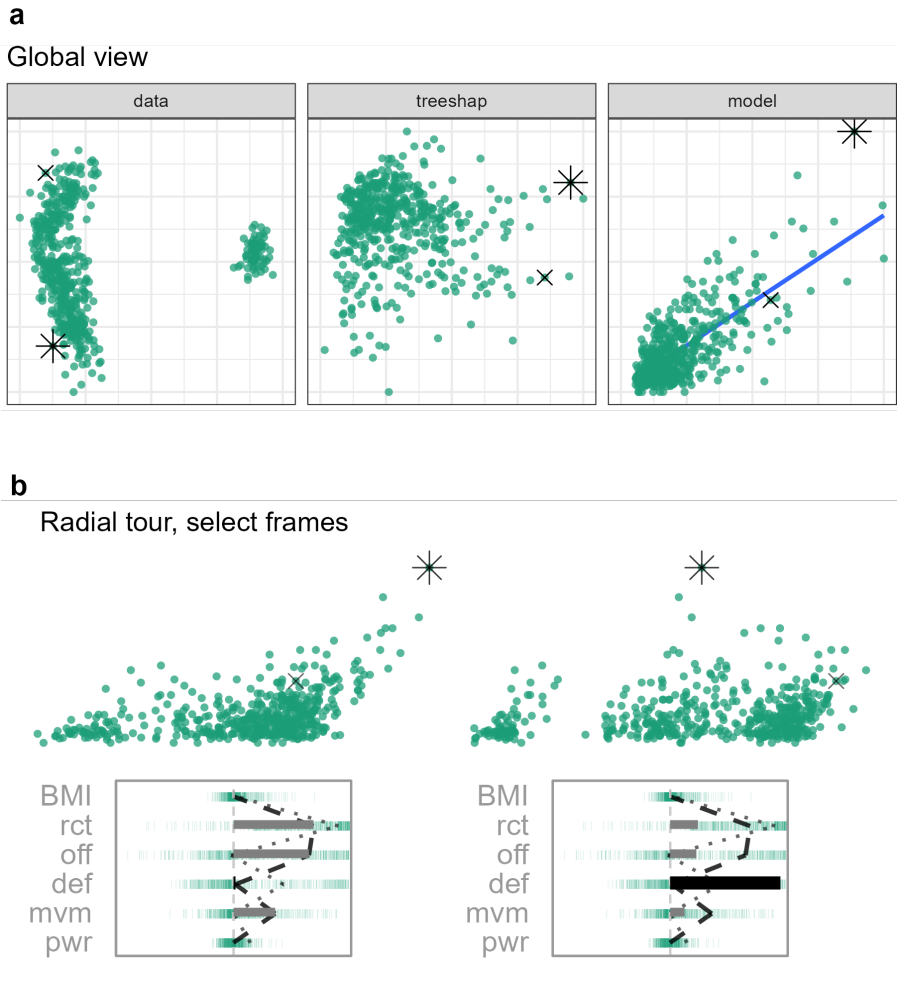

**Fig. 9** Exploring the wages relative to skill measurements in the FIFA 2020 data. Star offensive player (L. Messi) is the PI, and he is compared with a top defensive player (V. van Dijk): (a) global view, (b) observed values vs linear combination of predictors (predicted values). The attribution projection produces a view where Messi has very high predicted (and observed) wages. Defense ('def') is the chosen variable to vary. It starts very low, and Messi's predicted wages decrease dramatically as its contribution increases (right plot). The increased contribution in defense comes at the expense of offensive and reaction skills. The interpretation is that Messi's high wages are most attributable to his offensive and reaction skills, as initially provided by the LVA. (A video of the animated radial tour can be found at https://vimeo.com/666431163.)

This paper has provided additional interactive graphics tools to utilize LEs to explore and understand model predictions. Several diagnostic plots are provided to assist with understanding the sensitivity of a prediction to particular variables. A global view shows the data space, explanation space, and residual plot, to get an overview of the distribution of LEs across all observations. The user can interactively select observations to compare, contrast, and study further. The LE is converted into an LVA (linear projection) where the radial tour can be used to understand the prediction's sensitivity to a particular variable.

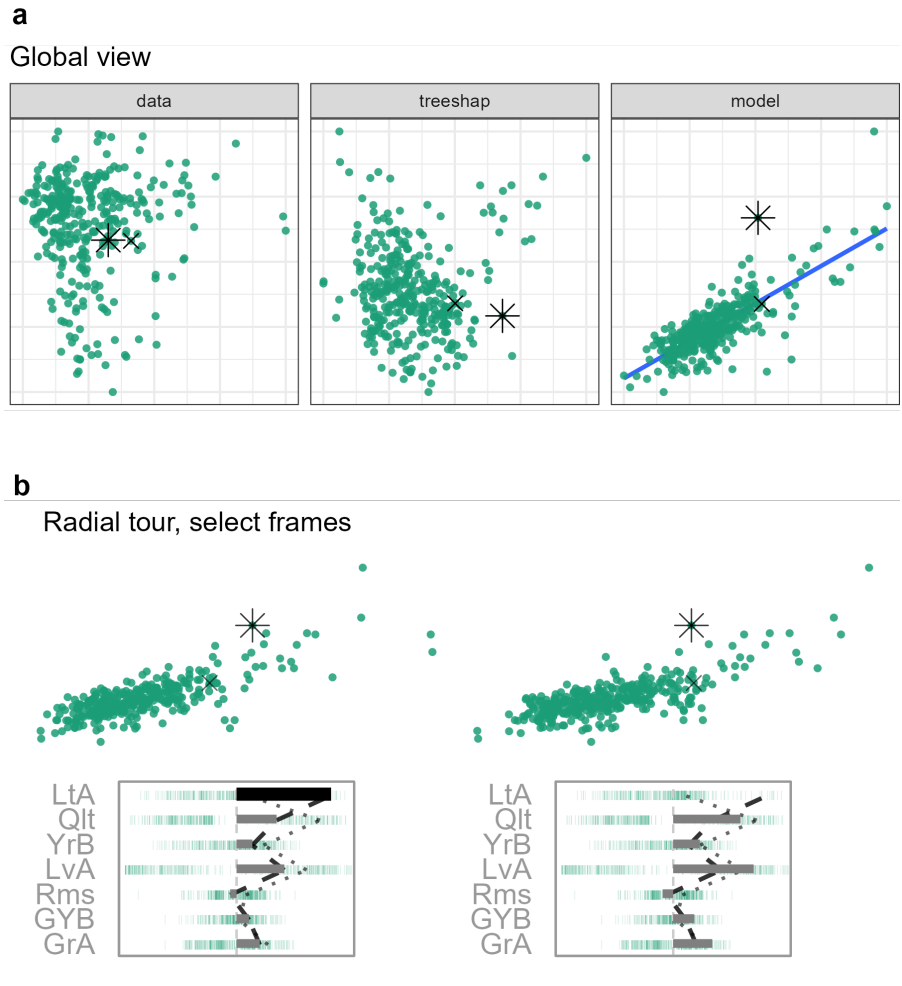

**Fig. 10** Exploring an observation with an extreme residual as the PI in relation to an observation with an accurate prediction for a similarly priced house in a random forest fit to the Ames housing data: (a) global view, (b) observed values vs linear combination of predictors (predicted values). The LVA indicates a sizable attribution to Lot Area ($LtA$), while the CI has minimal attribution to this variable. The PI has a higher predicted value than the CI in the attribution projection. Reducing the contribution of Lot Area brings these two prices in line. This suggests that if the model did not value Lot Area so highly for this observation, then the observed sales price would be quite similar. That is, the large residual in the model is due to a lack of factoring Lot Area into the prediction of PI's sales price. (A video showing the animation is at https://vimeo.com/666431134.)

This approach has been illustrated using four data examples of random forest models with the tree SHAP LVA. LEs focus on the model fit and help to dissect which variables are most responsible for the fitted value. They can also form the basis of learning how the model has got it wrong, when the observation is misclassified or has a large residual.

In the penguins example, we showed how the misclassification of a penguin arose due to it having an unusually small flipper size compared to others of its species. This was verified by making a follow-up plot of the data. The chocolates example shows how a dark chocolate was misclassified primarily due to its attribution to Fiber, and a milk chocolate was misclassified as dark due to its lowish Sodium value. In the FIFA example, we show how low Messi's salary would be if it depended on their defensive skill. In the Ames housing data, an inaccurate prediction for a house was likely due to the lot area not being effectively used by the random forest model.

This analysis is manually intensive and thus only feasible for investigating a few observations. The recommended approach is to investigate an observation where the model has not predicted accurately and compare it with an observation with similar predictor values where the model fitted well. The radial tour launches from the attribution projection to enable exploration of the sensitivity of the prediction to any variable. It can be helpful to make additional plots of the variables and responses to cross-check interpretations made from the cheem viewer. This methodology provides an additional tool in the box for studying model fitting.

These tools work better for smaller data, because being able to interact with the plots is necessary. XAI has been developed to tackle large data. To work with bigger data sets, would involve subsetting it after modeling and computing the LEs, to keep a representative sample of well-fitted observations, along with the observations that are especially interesting to investigate.

There are many additional future directions for this work. Primarily, development should make it easier to focus on what can be learned from the LEs, to be able to compare different versions, to flag or annotate values, and output of log the results of interactive analysis.

# 7 Package Infrastructure

An implementation is provided in the open-source **R** package **cheem**, available on CRAN Spyrison (2023). Example data sets are provided. You can upload your own data after model fitting and computing the LVAs. The LVAs need to be pre-computed, possibly using the `cheem_ls()` function, and saved as an `rds` file. Examples show how to do this for tree SHAP values, using **treeshap** (tree-based models from **gbm**, **lightgbm**, **randomForest**, **ranger**, or **xgboost** Greenwell et al. (2020); Shi et al. (2022); Liaw and Wiener (2002); Wright and Ziegler (2017); Chen et al. (2021), respectively). The SHAP and oscillation explanations could be easily added using `DALEX::explain()` (Biecek, 2018; Biecek and Burzykowski, 2021).

The application was made with **shiny** (Chang et al., 2021). The tour visual is built with **spinifex** (Spyrison and Cook, 2020). Both views are created first with **ggplot2** (Wickham, 2016) and then rendered as interactive `html` widgets with **plotly** (Sievert, 2020). **DALEX** (Biecek, 2018) and *Explanatory Model Analysis* (Biecek and Burzykowski, 2021) are helpful for understanding LEs and how to apply them.

The package can be installed from CRAN, and the application can be run using the following **R** code:

```
install.packages("cheem", dependencies = TRUE)
library("cheem")
run_app()
```

A version of the cheem application can be accessed at https://nicholas-spyrison.shinyapps.io/cheem/, the development version of the package is available at https://github.com/nspyrison/cheem, and documentation of the package can be found at https://nspyrison.github.io/cheem/.

**Acknowledgments.** Kim Marriott provided advice on many aspects of this work, especially on the explanations in the applications section. This research was supported by the Australian Government Research Training Program (RTP) scholarships. Thanks to Jieyang Chong for helping proofread this article. The namesake, Cheem, refers to a fictional race of humanoid trees from Doctor Who lore. **DALEX** pulls on from that universe, and we initially apply tree SHAP explanations specific to tree-based models.